\documentclass[10pt,twocolumn,letterpaper]{article}
\usepackage{cvpr}
\usepackage{times}
\usepackage{epsfig}
\usepackage[pagebackref=true,breaklinks=true,letterpaper=true,colorlinks,bookmarks=false]{hyperref}
\usepackage{cleveref}
\usepackage{graphicx}
\usepackage{amsmath}
\usepackage{amssymb}
\usepackage{textcomp}
\usepackage{subcaption}
\usepackage{enumitem}

\usepackage[pagebackref=true,breaklinks=true,letterpaper=true,colorlinks,bookmarks=false]{hyperref}

\cvprfinalcopy 

\def\cvprPaperID{24} 

\ifcvprfinal\fi
\begin{document}

\title{Real-time Hair Segmentation and Recoloring on Mobile GPUs}

\author{Andrei Tkachenka \qquad Gregory Karpiak\qquad Andrey Vakunov \qquad Yury Kartynnik\\ 
Artsiom Ablavatski \qquad Valentin Bazarevsky \qquad Siargey Pisarchyk\\
Google Research\\
1600 Amphitheatre Pkwy, Mountain View, CA 94043, USA\\
{\tt\small \{atkach, gkarpiak, vakunov, kartynnik, artsiom, bazarevsky, siargey\}@google.com}
}

\maketitle

\begin{abstract}
   We present a novel approach for neural network-based hair segmentation from a single  camera  input  specifically designed for real-time, mobile  application.  Our  relatively small neural network produces a high-quality hair segmentation mask that is well suited  for AR  effects, \eg virtual hair recoloring. The proposed model achieves real-time inference speed on mobile GPUs (30 - 100+ FPS, depending on the device) with high accuracy. We also propose a very realistic hair recoloring scheme. Our method has been deployed in major AR application and is used by millions of users.
\end{abstract}

\section{Introduction}

Video segmentation is a widely used technique that enables movie directors and video content creators to separate the foreground of a scene from the background, and treat them as two different visual layers. By modifying a layer (\eg via colorization, masking or replacement), creators and users can achieve compelling visual effects. However, this operation has traditionally been performed as a time-consuming manual process (\eg an artist rotoscoping every frame) or requires a studio environment with a green screen for real-time background removal (a technique referred to as chroma keying). In order to enable users to create this effect live in the viewfinder for AR applications, we designed a novel technique that is suitable for mobile phones. 

Specifically, we address the problem of computing a high fidelity hair segmentation mask from the video frame (without the use of a depth sensor) to enable very realistic hair recoloring effects. To achieve this, we leverage machine learning to solve a semantic segmentation task using convolutional neural networks. In particular, we designed a network architecture and training procedure suitable for mobile phones focusing on the following requirements and constraints:

\begin{itemize}
\setlength{\parskip}{0pt}
\setlength{\itemsep}{0pt plus 1pt}
    \item A mobile solution needs to be lightweight and ~10-30 times faster than existing state-of-the-art photo segmentation models. For real time performance, such a model needs to to be able to infer results at a minimum of 30 frames per second.
    \item A video model should leverage temporal redundancy (neighboring frames look similar) and exhibit temporal consistency (neighboring results should be similar)
\end{itemize}

\section{Network input}
\begin{figure}[tb]
    \centering
        \includegraphics[width=\linewidth]{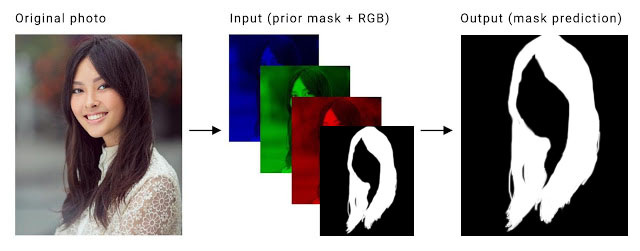}
    \caption{Network input for our segmentation model. To achieve temporal consistency we separate the original frame (left) in its three color channels and concatenate it with the previous mask (middle), yielding a 4 channel input tensor. This is used as input to our neural network to predict the mask for the current frame (right).}
    \label{fig:network_input}
\end{figure}

Our specific segmentation task is to compute a soft binary mask separating hairs from background for every input frame of the video (three channels, RGB, no depth input required). For AR applications, achieving temporal consistency of the computed masks across frames is key. Current methods utilize LSTMs~\cite{hochreiter1997long} to realize this but are too computationally expensive for real-time applications on mobile phones. Instead, we propose to pass the segmentation mask computed for for the previous frame as a prior by concatenating it as fourth channel to the current RGB input frame, as shown in \Cref{fig:network_input}.
The original frame (left) is separated in its three color channels and concatenated with the previous mask (middle). This is used as input to our neural network to predict the mask for the current frame (right).

\section{Dataset and ground truth annotation}
\begin{figure}[tb]
    \centering
        \includegraphics[width=\linewidth]{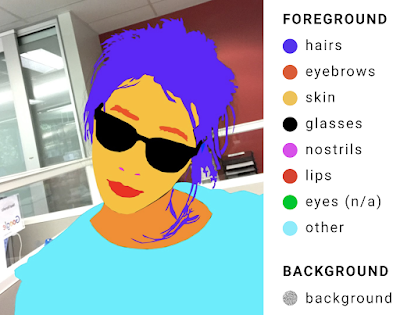}
    \caption{Dataset annotation example. See text for details.}
    \label{fig:dataset_example}
\end{figure}

High quality segmentation results require high quality, ground truth annotations. To provide high quality data for our machine learning pipeline, we annotated tens of thousands of images that captured a wide spectrum of human poses and background settings. One of the examples is shown in \Cref{fig:dataset_example}.  Annotations consists of pixel-accurate locations of foreground elements such as hair, glasses, neck, skin, lips, etc. Hair is one of the most difficult classes to mark up due to its thin structure. To quantify the difficulty of the task, we measured human annotator consistency by providing the same image to different annotators. The cross-validation across their hair markup was measured at 88\% Intersection-Over-Union (IOU). We consider this number to be the upper bound, when evaluating  a model for quality.

\section {Training Procedure}

 For high-quality video segmentation, besides achieving frame-to-frame temporal continuity, we also need to account for temporal discontinuities such as people suddenly appearing in the field of view of the camera. To handle these use cases robustly, we transform the annotated ground truth in several ways when using it as a previous frame mask:

\vspace{-0.5\topsep}
\begin{itemize}
\setlength{\parskip}{0pt}
\setlength{\itemsep}{0pt plus 1pt}
    \item Empty previous mask: Trains the network to work correctly for the first frame and new objects in scene. This emulates the case of someone appearing in the camera's frame.
    
    \item Affine transformed ground truth mask: Minor transformations train the network to propagate and adjust to the previous frame mask. Major transformations train the network to understand inadequate masks and discard them.
    
    \item Transformed image - we implement thin plate spline smoothing of the original image to emulate fast camera movements and rotations.
    
\end{itemize}
\vspace{-0.5\topsep}

\section{Network architecture}

\begin{figure}[tb]
    \centering
        \includegraphics[width=\linewidth]{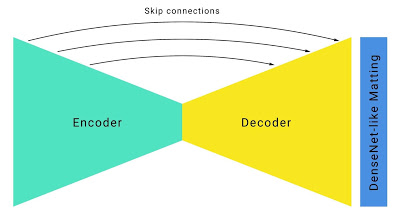}
    \caption{Network architecture, see text for details.}
    \label{fig:network_architecture}
\end{figure}

We build on a standard hourglass segmentation network architecture with skip connection~\cite{ronneberger2015u} shown in \Cref{fig:network_architecture}, customized for real-time mobile inference by adding the following improvements:
\vspace{-0.5\topsep}
\begin{itemize}
\setlength{\parskip}{0pt}
\setlength{\itemsep}{0pt plus 1pt}
    \item We employ big convolution kernels with large strides of four and above to detect object features on the high-resolution RGB input frame. Convolutions for layers with a small number of channels (as it is the case for the RGB input) are comparably cheap, so using big kernels here has almost no effect on the computational costs.

    \item For speed gains, we aggressively downsample using large strides combined with skip connections like U-Net to restore low-level features during upsampling. For our segmentation model this technique results in a significant improvement of 5\% IOU compared to using no skip connections.

    \item For even further speed gains, we optimized E-Net bottlenecks. We noticed that squeezing layers more aggressively~\cite{iandola2016squeezenet} by a factor of 16 or 32 can be done without significant quality degradation.

    \item To refine and improve the accuracy of edges, we added several DenseNet~\cite{huang2017densely} layers on top of our network in full resolution similar to neural matting. This technique improves overall model quality by a slight 0.5\% IOU, however perceptual quality of segmentation improves significantly.
\end{itemize}
\vspace{-0.5\topsep}

\section{Hair recoloring}

One compelling application for segmentation, and in particular for hair segmentation, is realistic hair recoloring. Given the original image and the computed hair segmentation mask, we can recolor hair in a realistic looking way, by carefully selecting a set of effects specifically tuned for the image and the kind of hair. However, the same set of effects might not necessarily work for a different kind of hair or a photo taken with a different device or settings. Hence, the recoloring procedure requires manual adjustments, which can be tedious and time consuming.

In order to automate the procedure and enable it for a wider range of possible inputs (one of our goals is to make recoloring work for all kinds of hair) to be able to run both segmentation and recoloring in real time, we propose the following two step technique:
\vspace{-0.5\topsep}
\paragraph{Preparation:}
\vspace{-0.5\topsep}
\begin{itemize}
\setlength{\parskip}{0pt}
\setlength{\itemsep}{0pt plus 1pt}
    \item Select two hair reference images: one having a very light hair color and another one having a very dark hair color.
    \item Calculate the average hair intensity for each reference image.
    \item Adjust and tune \textit{light} and \textit{dark} sets of effects producing desired hair color, when applied to light and dark reference images correspondingly. The procedure is manual and usually requires experimenting with a wide range of effects.
    \item Export the final sets of effects as reference lookup tables (LUTs) for further use.
\end{itemize}
\vspace{-0.5\topsep}


\vspace{-0.5\topsep}
\paragraph{Application:}
\vspace{-0.5\topsep}
\begin{itemize}
\setlength{\parskip}{0pt}
\setlength{\itemsep}{0pt plus 1pt}
    \item Obtain the hair segmentation mask for an arbitrary input image.
    \item Calculate the average hair intensity, given the input image and hair segmentation mask
    \item Interpolate the reference \textit{light} and \textit{dark} LUTs, using the corresponding pre-calculated intensity values as source and the above calculated hair intensity for the current frame as destination.
    \item Apply the interpolated LUT to the segmented hair area.
\end{itemize}
\vspace{-0.5\topsep}

\vspace{-0.5\topsep}

\begin{figure}[tb]
    \centering
        \includegraphics[width=\linewidth]{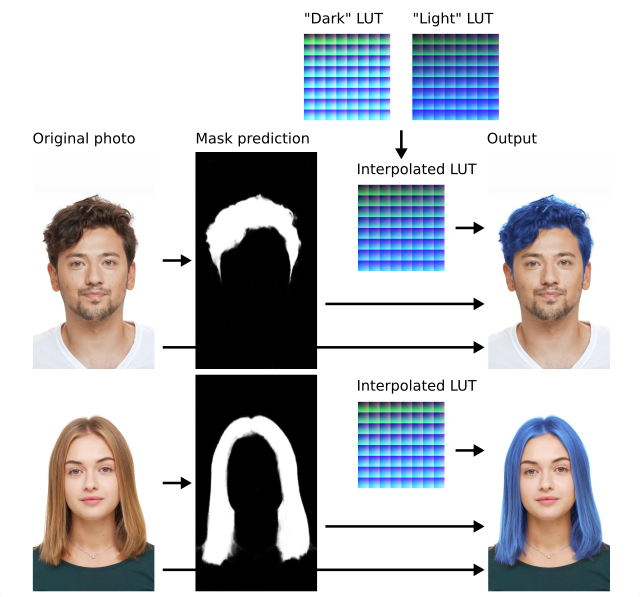}
    \caption{Hair recoloring procedure. For each hair color an artist designs the desired colorization effect for \emph{dark} and \emph{light} hair intensity, represented as LUTs. When applied to an image, the LUTs are interpolated to match the user's hair intensity. See text for details.}
    \label{fig:recoloring_scheme}
\end{figure}

\begin{table}[htb]
\centering
\begin{tabular}{|l|c|c|c|c|}
\hline
Model (input) & IOU & Time, ms & Time, ms \\
 & \% & (iPhone XS) & (Pixel 3) \\ \hline
 Full size (512$\times$512) & 81.0\% & 5.7 & 19 \\ \hline
 Small size (256$\times$256) & 80.2\% & 1.9 & 6 \\ \hline
\end{tabular}
\vskip 1ex
\caption{Model performance characteristics}
\label{tbl:inference_speed}
\end{table}

\section{Results and AR application}

We achieve real-time performance on a variety of mobile devices by designing our ML pipeline to leverage the mobile GPU end-to-end, from the camera input as GPU buffer, over GPU-based ML inference (using TFLite GPU \cite{TFLiteGPU}) and rendering (custom shaders for intensity calculation and LUT application). For inference numbers please refer to \cref{tbl:inference_speed}.

\begin{figure}[htb]
    \centering
        \includegraphics[width=\linewidth]{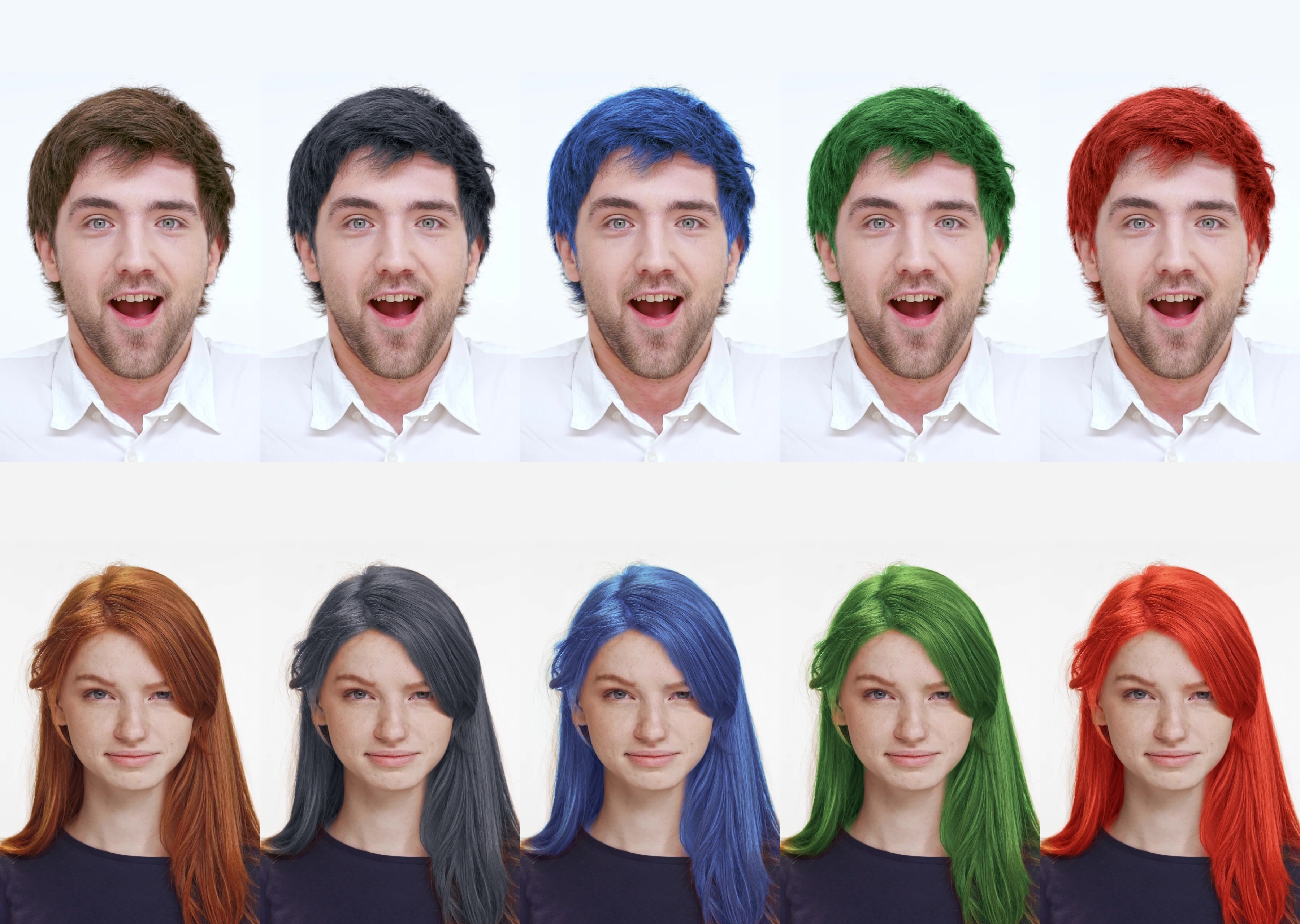}
    \caption{Hair recoloring samples using various hair colors. Note, how the recoloring is consistent across subjects, independent of the original hair color.}
    \label{fig:recoloring_samples}
\end{figure}

Our ML pipeline can be executed at higher than real-time speeds, achieving a realistic AR hair recoloring effect on mobile phones, as shown in \Cref{fig:recoloring_samples}.

{\small
\bibliographystyle{ieee_fullname}
\bibliography{segmentationbib}
}

\end{document}